# The risks of mixing dependency lengths from sequences of different length

Ramon Ferrer-i-Cancho

*Complexity and Quantitative Linguistics Lab, LARCA Research Group. Department of Computer Science, Universitat Politècnica de Catalunya (UPC). Campus Nord, Edifici Ω, Jordi Girona Salgado 1-3. 08034 Barcelona, Catalonia (Spain).*

E-mail: rferrericancho@lsi.upc.edu

Haitao Liu

*Department of Linguistics, Zhejiang University, No. 866 Yuhangtang Road, 310058, Hangzhou, China.*

E-mail: lhtzju@gmail.com

ABSTRACT

Mixing dependency lengths from sequences of different length is a common practice in language research. However, the empirical distribution of dependency lengths of sentences of the same length differs from that of sentences of varying length. The distribution of dependency lengths depends on sentence length for real sentences and also under the null hypothesis that dependencies connect vertices located in random positions of the sequence. This suggests that certain results, such as the distribution of syntactic dependency lengths mixing dependencies from sentences of varying length, could be a mere consequence of that mixing. Furthermore, differences in the global averages of dependency length (mixing lengths from sentences of varying length) for two different languages do not simply imply a priori that one language optimizes dependency lengths better than the other because those differences could be due to differences in the distribution of sentence lengths and other factors.

# 1. INTRODUCTION

The statistical properties of syntactic dependency lengths have been the subject of many studies over the last decade (Hiranuma 1999, Ferrer-i-Cancho 2004, Ferrer-i-Cancho

2006, Liu 2007, Gildea & Temperley 2007, Liu 2008, Temperley 2008, Gildea & Temperley 2010).

Here $p(d \mid n)$ is defined as the probability that a dependency has length $d$ in a sequence (e.g., a sentence) of length $n$, while $p(d)$ is defined as probability that a dependency has length $d$ regardless of the length of the sequence. $p(n)$ is defined as the probability that a sequence has length $n$. Then,

$$p(d) = \sum_{n=n_{min}}^{\infty} p(d \mid n) p(n), \qquad (1)$$

being $n_{min}$ the minimum sentence length (e.g., $n_{min} = 2$).

If $D$ is defined as the sum of the dependency lengths of a sequence then $\langle d \rangle = D/(n-1)$ is the mean dependency length of a sequence (assuming that dependencies form a tree and then there are $n - 1$ dependencies in a sentence of length $n$). $E[\langle d \rangle \mid n]$, the expected mean dependency length in sequences of length $n$, is defined as

$$E[\langle d \rangle \mid n] = E[d \mid n] = \sum_{d=1}^{n-1} p(d \mid n) d, \qquad (2)$$

while the expectation of $\langle d \rangle$ and $D$ over sentences of varying length are, respectively,

$$E[\langle d \rangle] = \sum_{n=n_{min}}^{\infty} p(n) E[\langle d \rangle \mid n] \qquad (3)$$

and

$$E[D] = \sum_{n=n_{min}}^{\infty} p(n) E[D \mid n]. \qquad (4)$$

As $E[\langle d \rangle \mid n] = E[d \mid n]$ (Ferrer-i-Cancho 2013), one has that $E[\langle d \rangle] = E[d]$ according to Eq. 3.

In research on various statistical aspects of syntactic dependency lengths, $E[\langle d \rangle \mid n]$ is estimated as the mean over mean dependency lengths of sentences of length $n$ words (e.g., Ferrer-i-Cancho 2004), $E[d]$ is estimated as the mean of $d$ over all the syntactic dependencies of a treebank (Liu 2008) and $E[D]$ is estimated by the mean of $D$ over all the sentences of a treebank (Gildea & Temperley 2010).

The estimated $E[\langle d \rangle | n]$ in syntactic dependencies is in-between the minimum possible and a random linear arrangement of vertices (Ferrer-i-Cancho 2004, Ferrer-i-Cancho 2006). While estimates of $E[\langle d \rangle | n]$ scale linearly with sentence length in a random linear arrangement of words, i.e. $E[\langle d \rangle | n] = (n+1)/3$, the estimated $E[\langle d \rangle | n]$ in real sentences scales sublinearly (Ferrer-i-Cancho 2004). Similarly, $E[d]$ and $E[D]$, are found to be between the minimum possible in different kinds of random control configurations (Gildea & Temperley 2007, Liu 2008, Temperley 2008, Gildea & Temperley 2010). This article analyzes the general problems of mixing dependency lengths from sequences of varying length in $p(d)$, $E[d]$ or $E[D]$, being the syntactic dependencies between the words word pairs of a sentence a particular case of application (Mel'čuk 1988, Hudson 2007).

## 2. THE PROBLEMS OF MIXING DEPENDENCY LENGTHS FROM SENTENCES OF DIFFERENT LENGTH

### 2.1. Empirical arguments.

The distribution of syntactic dependency lengths of sentences of a given length is not necessarily consistent with the distribution of mixed dependency lengths. An exponential distribution for $p(d | n)$ has been suggested focusing on sentences of a given length (Ferrer-i-Cancho 2004) while a right-truncated zeta distribution has been suggested for $p(d)$ (Liu 2007). However, both suggestions must be explored further. Concerning $p(d)$, it has only been investigated within small Chinese texts of lengths of 200-400 words (Liu 2007). Concerning $p(d | n)$, it has been investigated in much larger corpora but only for certain sentence lengths: $n = n^*$ being $n^*$ the typical sentence length or $n \approx \langle n \rangle$ being $\langle n \rangle$ the mean sentence length in a Czech and a Romanian treebank. Besides, the hypothetical exponential distribution seems to have two regimes with a breakpoint at distance d ≈ 5 in Czech which has not been sufficiently investigated. Future research should consider other languages and other sentence lengths. The possibility that the differences between *p(d | n)* and *p(d)* are simply due to typological differences between languages or differences in genre within a language cannot be denied.

Interestingly, the estimated $E[\langle d \rangle | n]$ scales sublinearly as a function of *n* in Basque, Catalan and Spanish (Fig. 1). This indicates that the distribution of dependency lengths of sentences with different lengths is not the same. A preliminary study in a Romanian collection of sentences (Ferrer-i-Cancho 2004) indicated that the growth of $E[\langle d \rangle | n]$ was very slow assuming a linear dependency between $E[\langle d \rangle | n]$ and *n*. However, the functional dependency between $E[\langle d \rangle | n]$ and *n* is not known and should be the subject of future research. Another question for further research is determining which of these two hypotheses is more appropriate:
1. The mathematical form of the distribution is the same for any sentence length but its parameters change depending on *n*.
2. The mathematical form of the function (not only the parameters), depend on *n*.

## 2.2. Theoretical arguments

Under the null hypothesis of dependencies being formed with pairs of vertices taking random positions of the sequence, the distance between linked vertices follows a decreasing linear distribution (Ferrer-i-Cancho 2004), i.e. the probability that an edge connects vertices at distance *d* is

$$p(d) = \frac{2(n-d)}{n(n-1)} \qquad (5)$$

with *p(d)* = 0 for *d* < 1 or *d* > *n* – 1.
Notice that the null distribution has one parameter, i.e. *n*, so *p(d)* depends on the length of the sentence. Under this null hypothesis, $E[\langle d \rangle | n] = E[d | n] = (n+1)/3$ (Ferrer-i-Cancho 2013).

Obviously, *d* is bounded above by *n* - 1. In general, the limits of the variation of $\langle d \rangle$ in a sentence (and thus those of $D = \langle d \rangle (n-1)$) depend on *n*, too. $\langle d \rangle_{min}$ and $\langle d \rangle_{max}$ are defined, respectively, as the minimum and maximum value of $\langle d \rangle$ that can be reached. Obviously, $\langle d \rangle_{max} \leq n - 1$ (Ferrer-i-Cancho 2013). In a non-crossing tree, $\langle d \rangle_{max} = n/2$ (Ferrer-i-Cancho 2013). As far as we know, $\langle d \rangle_{max}$ has not been investigated for trees where crossings are allowed. In general,

$$\langle d \rangle_{min} \geq \frac{n \langle k^2 \rangle}{8(n-1)} + \frac{1}{2}, \qquad (6)$$

where $\langle k^2 \rangle$ is the second moment about zero of the degree of the dependency tree (Ferrer-i-Cancho 2013). The dependency with $n$ is obvious but a priori it cannot be excluded for $\langle k^2 \rangle$, which is bounded below by its value in a linear tree and bounded above by its value in a star tree (Ferrer-i-Cancho 2013), i.e.

$$4 - \frac{6}{n} \leq \langle k^2 \rangle \leq n - 1. \qquad (7)$$

However, the relationship between $\langle k^2 \rangle$ and $n$ in real sentences should be investigated.

An exponential distribution for $p(d \mid n)$ has been derived mathematically using language independent cognitive pressures (Ferrer-i-Cancho 2004) but the empirical distribution suggests two exponential regimes that are not covered by that simple distribution and have not been explained to our knowledge. Further research should be performed to determine if the shape of $p(d \mid n)$ depends on certain variables such as the type of language or genre as suggested by quantitative research on dependency lengths (Liu 2008).

The fact that a zeta distribution has been proposed for $p(d)$ while an exponential distribution has been proposed for $p(d \mid n)$ (using both empirical and theoretical arguments) suggests that $p(d)$ may not be theoretically informative. One possibility is that $p(d)$ is a trivial consequence of mixing exponentially distributed variables with different parameters. Indeed, a power-law distribution can arise aggregating information that is not power-law distributed in different fashions (Stumpf & Porter 2012). Specially relevant here is the emergence of power-law distributions by combining elements of different types which have varying distributions (Tanaka et al 2005). Power-laws can be reproduced by a superposition of other distributions, for instance, exponential distributions (Popescu et al 2009). Eq. 1 suggests a possible track for compounding in dependency lengths.

Another statistical caveat is the possibility that two treebanks A and B satisfy $E_A[d] > E_B[d]$ which can be prematurely interpreted as unequivocal evidence that the dependency lengths of B are more optimized than those of A. However, $E_A[d] > E_B[d]$ does not exclude that neither A nor B is optimizing dependency lengths within sentences. Under the null hypothesis that the vertices of the dependency network are placed at random in a sequence (i.e. no dependency length minimization at all) and that the sequence length is at least two ($n \geq 2$), one has (Appendix B)

$$E[d] = \frac{1}{3}(E[n] + 1), \quad (8)$$

where $E[n]$ is the expectation of the sequence length. Therefore $E[n]$ determines $E[d]$. Accordingly, Liu (2008) found that estimates of $E[d]$ of random controls are more strongly correlated with the mean sentence length (what he called MSL) than estimates of $E[d]$ from real sentences: Eq. 8 indicates that the mean dependency length is a perfect function of the mean sentence length under the null hypothesis.

Imagine that sequence lengths go from 2 to $n_{max}$ and that $n$ is distributed uniformly in treebank A. Then (Appendix B)

$$E_A[n] = \frac{1}{n_{max} - 1}\left(\frac{n_{max}(n_{max} + 1)}{2} - 1\right). \quad (9)$$

Imagine next that sequence lengths vary in the same interval and that $n$ is distributed by a kind of truncated zeta distribution, then (Appendix B)

$$E_B[n] = \frac{n_{max} - 1}{\sum_{n=2}^{n_{max}} 1/n}. \quad (10)$$

Fig. 2 shows that $E_A[n] > E_B[n]$ for $n_{max} > 2$ and thus, according to Eq. 8, $E_A[d] > E_B[d]$ in the same range, but this does not imply that A is optimizing dependency lengths better than B within sentences: A and B are both placing vertices within a sentence "at random" in colloquial terms. However, the distribution of sequence lengths might be responsible for some degree of optimization, but not one that impacts on dependency lengths of sequences of the same length below chance.

The same problem of concerns E[$D$], which under the null hypothesis of random vertex placement becomes (Appendix B)

$$E[D] = \frac{1}{3}(E[n^2] - 1). \tag{11}$$

Eq. 11 indicates that E[$D$] is also determined by the distribution of sequence lengths under the null hypothesis and hence two treebanks A and B may satisfy $E_A[D] > E_B[D]$ but this does not mean that treebank B is more optimized within sentences.

An apparently little problem which has not been addressed when estimating E[$d$] or E[$D$] to our knowledge is the suitable value of $n_{min}$ (recall Eqs. 3 and 4). If one wanted to show that E[$d$] or E[$D$] are being minimized or decide which of two languages is more optimized based upon any of those global metrics, sentences where no optimization can be performed should be excluded. One may argue that $n_{min}=2$ because sequences of length 0 or 1 cannot have dependencies (in Appendix B we have assumed $n_{min}=2$ as this is needed by the formula for E[$d \mid n$] under the null hypothesis). However, notice that all the possible orderings of the vertices yield the same $D$ when $n = 2$ (Ferrer-i-Cancho 2008). Thus, $n_{min}=3$ might be more convenient as this is the minimum value of $n$ needed so that the value of $d$ (or $D$) is neither unique nor undefined. To reduce confounds, it is convenient to not include sequences with less than three elements in mixtures of dependency length information from sentences of varying length.

## 3. DISCUSSION

Our arguments have implications for research on dependency treebanks. Liu (2007) studied various aspects of the distribution of dependency distances in a Chinese dependency treebank by mixing the distances coming from sentences of different lengths. As expected from our concerns, the distribution of dependency distances in the mixed sentence length study of Chinese sentences does not decay exponentially as in the case of the sentences of the same length in Czech and Romanian (Ferrer-i-Cancho 2004). However, the results are not fully comparable and should be controlled for language, genre and even maybe treebank size. That demonstrates the need of further research applying the same methods to a sample of languages as broad as possible.

The issue of mixing of sequence lengths also concerns the analysis of dependency lengths by means of global metrics of dependency length such as estimates of E[$d$] (Liu 2008) or E[$D$] (Gildea & Temperley 2007, Temperley 2008, Gildea & Temperley 2010).

Liu (2008) considered treebanks from 20 different languages and sorted them by E[$d$] and found that Chinese had the largest E[$d$] among them. Gildea and Temperly (2010) confirmed a previous finding by Liu, namely that German had "longer dependencies" than 17 other languages (including English) but employing E[$D$] instead of E[$d$] as Liu did. However, such difference does not imply that German is less optimized than English: take A as the German treebank and B as the treebank of English and apply the arguments in section 2. Indeed, the relative ordering of languages by E[$d$] or E[$D$] could be simply due differences in the distribution of sentence lengths $p(n)$ among various factors. Reaching a strong conclusion on one language being more optimized than another would require controlling for the genre or style making the treebank, as the distribution of sentence lengths is known to depend on the characteristics of an author (e.g., Yule 1939, Williams 1940, Sichel 1974). However, one cannot exclude the possibility that dependency length minimization plays an important role in the distribution of sentence lengths as we have reviewed above arguments showing that the variation of $\langle d \rangle$ (or equivalently $D = \langle d \rangle (n-1)$) depends on $n$ (e.g., Eq. 6). Furthermore, the reverse might also be possible, i.e. sentence length might play a relevant role for dependency length minimization. Indeed, the range of variation of dependency lengths depends on $n$ (Ferrer-i-Cancho 2013). The optimization of the cost of sentences may involve the tuning of both sentence lengths and the internal dependency lengths. The point is that differences in E[$d$] or E[$D$] between two languages do not imply differences in the degree of optimization of dependency lengths within sentences.

To avoid all the problems reviewed so far, it is customary to consider dependency distances as a function of the sentence length, $E[\langle d \rangle | n]$ or $E[D | n]$ for both theoretical and empirical research (Ferrer-i-Cancho 2004, Ferrer-i-Cancho 2006, Ferrer-i-Cancho 2008, Park & Levy 2009).

The problem of mixing in global measures is a recurring problem in the history of science. A recent examples comes from complex networks research: physicists tried to summarize correlations between the degrees of nodes making an edge using an intraclass correlation coefficient (Newman, 2002). Interestingly, they realized soon that such coefficient mixed heterogeneous information (e.g., nodes with radically different degree) and then decided to consider the scaling of the mean of nodes adjacent to a target node as a function of the degree of the target node to have a better picture of degree correlations (Serrano et al 2007). For instance, E[$d$] or E[$D$] might be significantly small but $E[d|n]$ or $E[D|n]$ may not be significantly small for certain values of $n$. Besides, a language A may have greater E[$d$] or E[$D$] than another language B but then $E[d|n]$ or $E[D|n]$ be smaller in A than in B for certain lengths.

Let us assume that the hypothesis that dependency length is being minimized or constrained in a statistically detectable fashion is correct (e.g., Ferrer-i-Cancho 2004, Ferrer-i-Cancho 2006, Liu 2008) and that $\langle k^2 \rangle$, the degree 2$^{nd}$ moment about zero, plays a crucial role concerning the minimum value of $\langle d \rangle$ or $D$ that can be achieved. Then, restricting the analysis to dependencies from sentences of the same length (Ferrer-i-Cancho 2004) might not warrant a sufficiently homogenous sample of dependency lengths: $\langle k^2 \rangle$ may also be relevant (recall Eq. 6). Dependency lengths are still insufficiently understood. Investigating the distribution of dependency lengths in sentences of the same length or how $\langle d \rangle$, $D$ or $\langle k^2 \rangle$ scale as a function of sentence length in a large sample of languages are urgent research questions. We hope that our considerations stimulate further research.

ACKNOWLEDGEMENTS

This manuscript derives from a much longer manuscript finished in 2007. We are grateful to an anonymous reviewer for his very valuable comments. We thank M. A. Martí for the opportunity to use the Catalan and the Spanish dependency treebanks. We are grateful to A. Díaz, I. Aldeazabal and I. Aduriz for the opportunity to use the Basque dependency treebank. This work was supported by the grant "Iniciació i reincorporació a la recerca" from the Universitat Politècnica de Catalunya and the grants BASMATI (TIN2011-27479-C04-03) and OpenMT-2 (TIN2009-14675-C03) from the Spanish Ministry of Science and Innovation for RFC and the grant "Quantitative Linguistic Research of Contemporary Chinese" (11&ZD188) from the National Social Science Foundation of China for HL.

# APPENDIX A: Global measures of dependency length.

Liu (2008) defined the mean dependency distance (MDD) of a treebank as

$$MDD = \frac{1}{N-s} \sum_{i=1}^{N-s} |DD_i|, \tag{A1}$$

where $N$ is the number of words of the treebank, $s$ is the number of sentences and $DD_i$ is the dependency distance (the difference of the positions of the dependent and its governor) of the $i$-th dependency. Assuming that the syntactic dependency structures of the sentences are trees, a sentence of length $n$ words contributes with $n$-1 dependencies, and thus the total number of dependencies of a treebank containing $N$ words in $s$ sentences is $N$-$s$.

We define $f(n,d)$ as the number of dependencies of length $d$ in the sentences of length $n$ of the treebank ($f(n,d)=0$ if $d<1$ or $d \geq n$). Thus MDD can be redefined as

$$MDD = \frac{1}{N-s} \sum_{n=n_{min}}^{\infty} \sum_{d=1}^{\infty} f(n,d)d, \tag{A2}$$

The fact that

$$N-s = \sum_{n=n_{min}}^{\infty} \sum_{d=1}^{\infty} f(n,d), \tag{A3}$$

allows one to define MDD in terms of the relative frequency $p(n,d)= f(n,d)/(N-s)$. i.e.

$$MDD = \sum_{n=n_{min}}^{\infty} \sum_{d=1}^{\infty} p(n,d)d = \sum_{d=1}^{\infty} d \sum_{n=n_{min}}^{\infty} p(n,d) = \sum_{d=1}^{\infty} p(d)d = E[d]. \tag{A4}$$

That is, MDD estimates E[$d$], the expectation of $d$.

Gildea & Temperley (2010) employed the average dependency length (ADL), which they calculated "by averaging the dependency lengths for each sentence". In our notation, those researchers computed the mean of $D$, the sum of dependency lengths of a sentence, over the ensemble of sentences of a treebank, i.e.

$$ADL = \frac{1}{s} \sum_{D=n_{min}-1}^{\infty} f(D)D, \quad (A5)$$

where $s$ is the number of sentences and $f(D)$ is the number of sentences with $D$ as the sum of dependency lengths. Notice that the summation in Eq. A5 starts at $D = n_{min} - 1$ because $D \geq n - 1$ ("a dependency between adjacent words has a length of 1" according to Gildea & Temperley (2010) and a sentence of length $n$ has $n$-1 dependencies assuming that the dependency structure is a tree).

## APPENDIX B: E[$d$] under two sequence length distributions.

The expectation of E[$d$] can be written as

$$E[d] = \sum_{n=n_{min}}^{n_{max}} p(n)E[d \mid n], \quad (B1)$$

where $p(n) = 0$ if $n < n_{min}$ or $n > n_{max}$. Assuming (a) $n_{min} = 2$ and (b) that dependents are arranged at random in sequences and thus $E[d \mid n] = (n+1)/3$ (Ferrer-i-Cancho 2004, Ferrer-i-Cancho, 2013), Eq. B1 becomes

$$E[d] = \frac{1}{3}\sum_{n=2}^{n_{max}} p(n)(n+1) = \frac{1}{3}\left(\sum_{n=2}^{n_{max}} p(n)n + \sum_{n=2}^{n_{max}} p(n)\right) = \frac{1}{3}(E[n]+1), \quad (B2)$$

where

$$E[n] = \sum_{n=2}^{n_{max}} p(n)n. \quad (B3)$$

As E[$d$] is determined by E[$n$] in that case, two different distributions for sequence length will be considered next: a uniform distribution and a kind of truncated zeta distribution, both satisfying $p(n) = 0$ if and only if $n < 2$ or $n > n_{max}$. The uniform distribution, namely $p(2)=p(3)=\ldots =p(n_{max})=1/(n_{max} - 1)$, transforms Eq. B3 into

$$E[n] = \frac{1}{n_{\max}-1}\sum_{n=2}^{n_{\max}} n = \frac{1}{n_{\max}-1}\left(\frac{n_{\max}(n_{\max}+1)}{2}-1\right). \tag{B4}$$

The truncated zeta distribution, given by

$$p(n) = \frac{1/n}{\sum_{n=2}^{n_{\max}} 1/n}, \tag{B5}$$

transforms Eq. 3 into

$$E[n] = \frac{1}{\sum_{n=2}^{n_{\max}} 1/n}\sum_{n=2}^{n_{\max}} 1 = \frac{n_{\max}-1}{\sum_{n=2}^{n_{\max}} 1/n}. \tag{B6}$$

Notice that Eq. B5 defines a left and right truncation ($p(n)=0$ for $n < 2$ or $n > n_{max}$) with regard to a standard zeta distribution (Wimmer & Altmann, 1999), where $p(n) > 0$ for $n > 0$ and finite $n$).

Similarly, the expectation of $E[D]$ can be written as

$$E[D] = \sum_{n=n_{\min}}^{n_{\max}} p(n)E[D\mid n] = \sum_{n=n_{\min}}^{n_{\max}} p(n)(n-1)E[\langle d\rangle \mid n] = \sum_{n=n_{\min}}^{n_{\max}} p(n)(n-1)E[d\mid n]. \tag{B7}$$

As we did for $E[d]$, assuming that vertices are ordered at random with $n_{min} = 2$, Eq. B7 becomes

$$E[D] = \frac{1}{3}\sum_{n=2}^{n_{\max}} p(n)(n-1)(n+1) = \frac{1}{3}\left(\sum_{n=2}^{n_{\max}} p(n)n^2 - \sum_{n=2}^{n_{\max}} p(n)\right) = \frac{1}{3}(E[n^2]-1). \tag{B8}$$

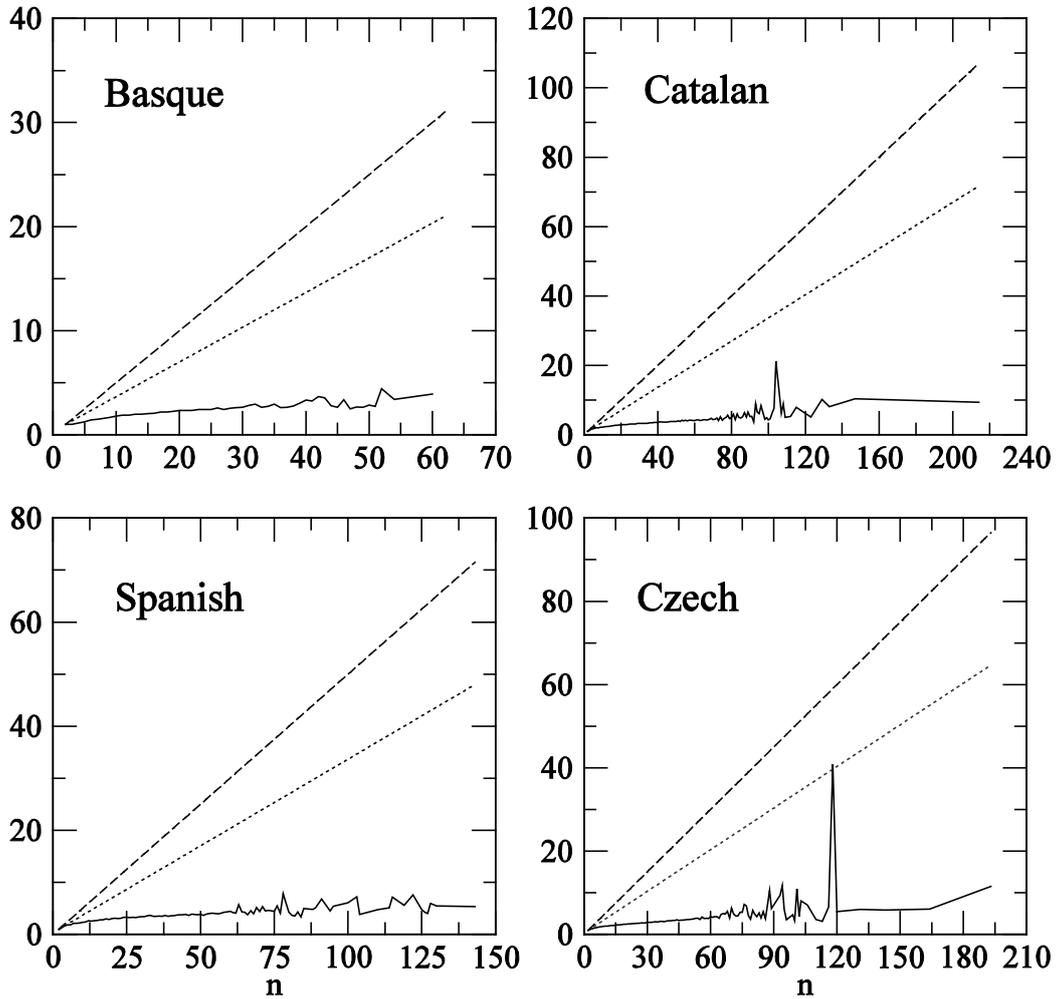

Figure 1. $\langle d \rangle$, the mean syntactic dependency length as a function of *n*, the sentence length (solid line) in the Basque dependency treebank Eus3LB (Aduriz *et al.* 2003, Palomar *et al.* 2004), the Catalan dependency treebank AnCora-Dep-CA, the Spanish dependency treebank AnCora-Dep-ES (Civit *et al.* 2006, Peris *et al.* 2010, Recasens & Martí 2010) and the Prague dependency treebank 1.0 for Czech (Böhmová *et al.* 2003). AnCora-Dep-CA and AnCora-Dep-ES freely available for research from http://clic.ub.edu/ancora/. For comparison, $E[\langle d \rangle]=(n+1)/3$, the expected value of $\langle d \rangle$ under random vertex placement (dotted line), and $\langle d \rangle_{max} = n/2$, the maximum value of $\langle d \rangle$ when crossings are not allowed (dashed line), are also shown. Sentences whose syntactic dependency graph is not a tree were excluded.

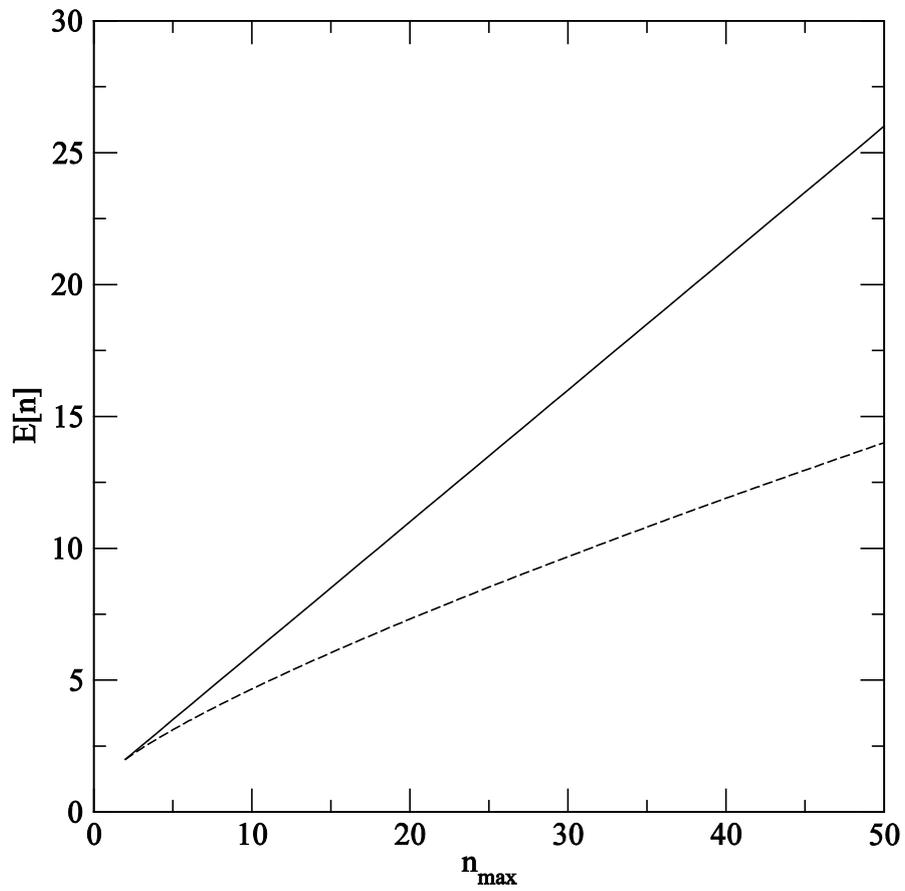

Figure 2. E[*n*], the expectation of sequence length, versus $n_{max}$, the maximum sequence length. Two distributions with sequence lengths between 2 and $n_{max}$ are considered: a uniform distribution (solid line) and a truncated zeta distribution (dashed line).